\documentclass[10pt, a4paper]{article}

\usepackage[final]{lrec2026} 
\usepackage{linguex}
\usepackage{graphicx}
\usepackage{xcolor}
\usepackage{color}
\usepackage{tabularx}

\title{Frame Semantic Patterns for Identifying Underreporting of Notifiable Events in Healthcare: The Case of Gender-Based Violence}

\name{Lívia Dutra\textsuperscript{1,2}, Arthur Lorenzi\textsuperscript{1,3}, Laís Berno\textsuperscript{1}, Franciany Campos\textsuperscript{1},\\
{\bf \large Karoline Biscardi\textsuperscript{1,4}, Kenneth Brown\textsuperscript{1}, Marcelo Viridiano\textsuperscript{1}, Frederico Belcavello\textsuperscript{1},}\\
{\bf \large Ely Matos\textsuperscript{1}, Olívia Guaranha\textsuperscript{3}, Erik Santos\textsuperscript{3}, Sofia Reinach\textsuperscript{3}, Tiago Timponi Torrent\textsuperscript{1,5}}} 

\address{\textsuperscript{1} Federal University of Juiz de Fora | FrameNet Brasil, \textsuperscript{2} University of Gothenburg, \\ 
\textsuperscript{3} Vital Strategies Brasil,
\textsuperscript{4} Federal University of Minas Gerais, \\
\textsuperscript{5} Brazilian National Council for Scientific and Technological Development – CNPq \\
         livia.vicente.dutra@svenska.gu.se, \{alorenzi, oguaranha, esantos, sreinach\}@vitalstrategies.org, \\ 
         \{lais.berno, franciany.campos, kenneth.cyrill, marcelo.viridiano\}@estudante.ufjf.br, \\
         karolbiscardi@let.grad.ufmg.br, \{fred.belcavello, ely.matos, tiago.torrent\}@ufjf.br\\}

\abstract{
We introduce a methodology for the identification of notifiable events in the domain of healthcare. The methodology harnesses semantic frames to define fine-grained patterns and search them in unstructured data, namely, open-text fields in e-medical records. We apply the methodology to the problem of underreporting of gender-based violence (GBV) in e-medical records produced during patients' visits to primary care units. A total of eight patterns are defined and searched on a corpus of 21 million sentences in Brazilian Portuguese extracted from e-SUS APS. The results are manually evaluated by linguists and the precision of each pattern measured. Our findings reveal that the methodology effectively identifies reports of violence with a precision of 0.726, confirming its robustness. Designed as a transparent, efficient, low-carbon, and language-agnostic pipeline, the approach can be easily adapted to other health surveillance contexts, contributing to the broader, ethical, and explainable use of NLP in public health systems.
 \\ \newline \Keywords{FrameNet, Digital Healthcare Surveillance, Gender-Based Violence} }

\begin{document}

\maketitleabstract

\section{Introduction}

Healthcare surveillance is a major challenge for health professionals, even in the context of the growing digitalization of medical records and the widespread use of advanced tools for data analysis. In a review paper on the topic, \citet{10.1093/jamia/ocaa186} point out, among the major challenges in implementing adequate surveillance, the difficulty in analyzing unstructured data, the low quality of information in e-medical records, as well as privacy and security issues. \citet{nazi-peng-2024}, in turn, list nine challenges for the use of LLMs in healthcare, which include: in the dataset gathering phase, (a) presence of Personally Identifiable Information (PII), (b) possible privacy/security breaches and (c) bias introduced by uneven data distributions; in the model training phase, (d) lack of access to the workings of blackbox models and (e) carbon footprint; and, finally, in the application phase, lack of (f) transparency and (g) explainability, (h) ethical risks and (i) hallucinations.   

\begin{figure}[!ht]
    \centering
    \includegraphics[width=1\linewidth]{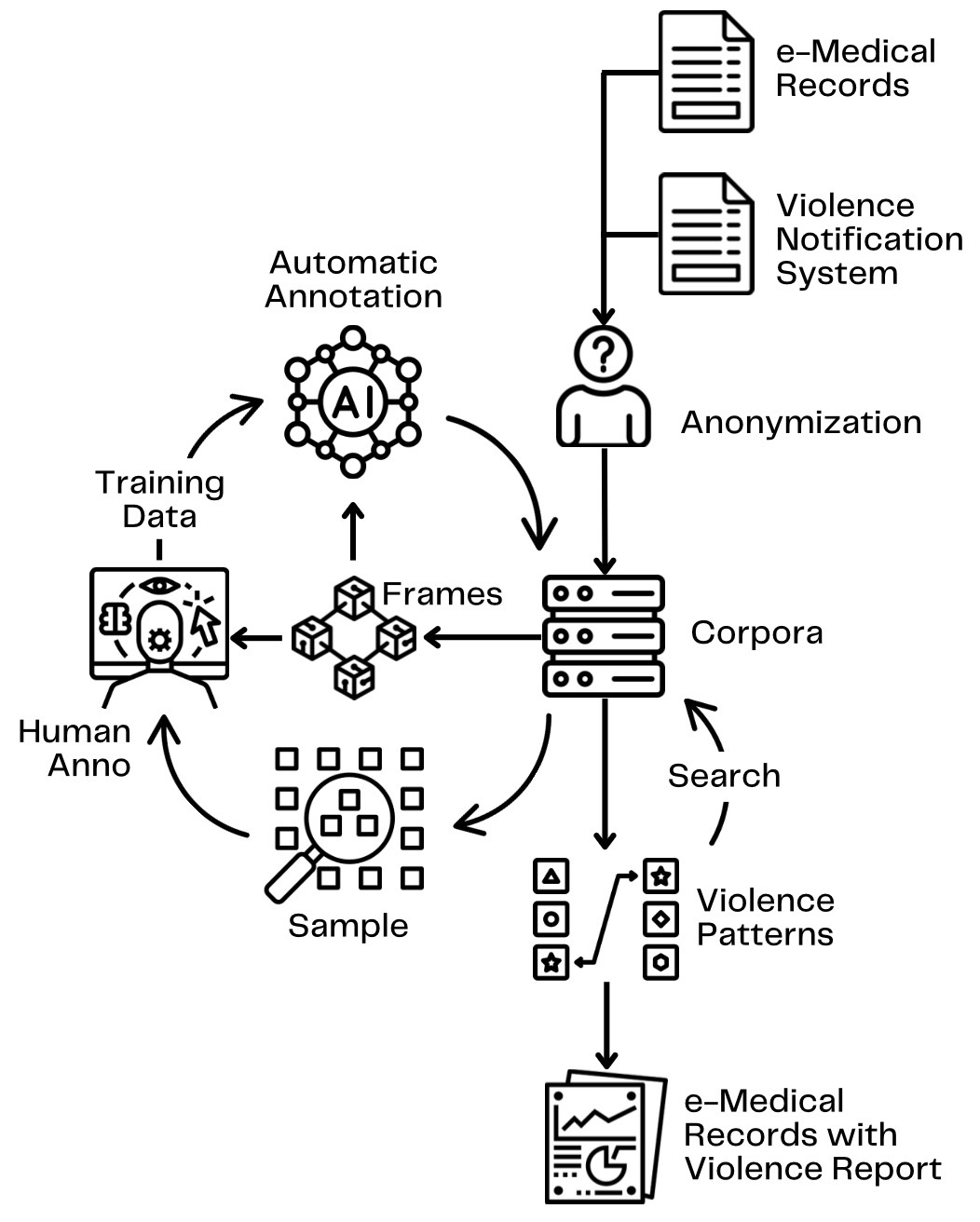}
    \caption{Schematization of the methodology}
    \label{fig:pipeline}
\end{figure}

Within the context of surveillance, Gender-Based Violence (GBV) remains a persistent challenge worldwide, affecting millions of women and girls in different cultural and social contexts. Global surveys suggest that more than one in three women who have ever been married or partnered have experienced physical or sexual violence from an intimate partner \citep{WHO2024}. Various studies treat Gender-Based Violence (GBV) as a public health issue \citep{garcia2011violence,sweet2014every,ohman2020public},  highlighting the importance of the healthcare system in the fight against GBV. In Brazil, underreporting is still a major issue when it comes to GBV cases. According to \citet{saliba2015desafios} and \citet{kind2013subnotificacao}, underreporting is usually due to excessive workload of healthcare teams, lack of knowledge about the importance of notification, fear of retaliation, as well as difficulties in identifying which injuries and conditions are related to an episode of violence.

This paper presents an innovative methodology, outlined in Figure \ref{fig:pipeline}, to improve healthcare surveillance in general and to tackle GBV underreporting specifically, by leveraging the modeling of frames covering the domains of healthcare \citep{dutra-et-al-2023} and violence \citep{larre-torrent-2024} in Brazilian Portuguese to identify GBV patterns in open text fields of e-medical records. We present the steps involved in the modeling and pattern definition process and evaluate the precision of the proposed methodology against a sample of 5,000 anonymized sentences extracted from e-medical records of primary care visits. We also conduct an error analysis and discuss its impact in feeding back the pattern-definition effort. Results show a precision of 0.726 of the proposed method in identifying GBV and 0.600 in the fine-grained classification of the type of violence.

Our contributions are threefold: (i) we introduce an innovative, explainable, and transparent pipeline for the identification of health events in e-medical records and report its energy consumption and carbon footprint; (ii) we evaluate the methodology for the challenging context of GBV in healthcare surveillance; and (iii) we design the methodology to be generalizable to healthcare surveillance needs beyond GBV. 

In the remainder of this paper we briefly introduce FrameNet, the framework under which the domains of healthcare and violence were modeled, and describe the modeling process employed for the two domains in Section \ref{sec:modeling}. Next, Section \ref{sec:anno} shows how the manual and automatic annotation of the open-text fields in the e-medical records was carried out. Section \ref{sec:pattern} focuses on the construction of frame-semantic patterns and on the methods used for their initial validation. Section \ref{sec:experiment} presents the setup used to evaluate the methodology, while Section \ref{sec:results} presents and discusses the results. 

\section{Frame-Based Modeling of the Healthcare and Violence Domains}
\label{sec:modeling}

Developed by Charles Fillmore in the late 1970s, Frame Semantics is based on the idea that the meaning of a particular lexical item is understood against the evocation of an associated conceptual scene. For its part, a frame is a structure of related concepts organized around a specific context in which certain words function as shortcuts to such a frame. That is, by using them, participants in a conversation not only instantly access the shared knowledge of that situation, but also classify elements within it, assuming that the general context is already understood by all involved \citep{Fillmore1982}.

As an implementation of Frame Semantics, Berkeley FrameNet (BFN) is a resource developed in 1997 to analyze the English lexicon. Since the inception of this project, the FrameNet model has been adapted to create similar resources for multiple other languages, such as Brazilian Portuguese: the FrameNet Brasil (FN-Br) \citep{torrent-et-al-2022}. The core principle of any framenet analysis is that "meaning is relativized to scenes" \citep{TheCaseforCaseReopened}. Simply put, the meaning of any word is determined by a background situation defined by the participants and props involved in it. Therefore, framenets, which are structured according to the principles of Frame Semantics, theoretically have the capacity to computationally represent diverse aspects of context \citep{torrent-et-al-2022}.

Every frame consists of a definition and a set of Frame Elements (FEs). The definition provides a general description of the scene the frame represents, typically referencing the FEs. These FEs, in turn, are classified as core FEs, mandatory to the frame's meaning, and non-core FEs, encoding circumstantial information. To illustrate the structure of a frame, consider the \texttt{Experience\_bodily\_harm} frame in \ref{ex:frame}, which models a scenario of physical injury. 

\ex.\label{ex:frame}\textbf{\texttt{Experience\_bodily\_harm}}\\
\textbf{Definition}: An \colorbox{blue}{\textcolor{white}{Experiencer}} is involved in a bodily injury to a \colorbox{violet}{\textcolor{white}{Body\_part}}. Often an \colorbox{gray}{\textcolor{white}{Injuring\_entity}} on which the \colorbox{blue}{\textcolor{white}{Experiencer}} injures themselves is mentioned.\\
\textbf{Core Frame Elements}:\\
\colorbox{violet}{\textcolor{white}{Body\_part}}: The location on the body of the \colorbox{blue}{\textcolor{white}{Experiencer}} where the bodily injury takes place.\\
\colorbox{blue}{\textcolor{white}{Experiencer}}: The being or entity that is injured.\\
\textbf{Peripheral Frame Elements}:\\
\colorbox{gray}{\textcolor{white}{Injuring\_entity}}: The \colorbox{blue}{\textcolor{white}{Experiencer}} injures her/himself on an \colorbox{gray}{\textcolor{white}{Injuring\_entity}}.\\
\colorbox{red}{\textcolor{white}{Severity}}: The extent to which the \colorbox{blue}{\textcolor{white}{Experiencer}} is affected by the injury.\\
\textbf{Frame-to-Frame Relations}:\\
is inchoative of: \texttt{Being\_hurt}\\
inherits from: \texttt{Undergoing}\\
is used by: \texttt{Cause\_harm}\\
\textbf{Lexical Units}: \textit{injury.n; break.v; bruise.v...}

Note that the definition mentions the main elements involved in the frame: the \textsc{Experiencer}, the \textsc{Body\_part} and the \textsc{Injuring\_entity}, the two first being core, and the last being peripheral, since this frame assumes the perspective of the person suffering the injury. Each frame element receives a definition that is specific to the frame. Moreover, because FrameNet is a net, not a list of frames, relations between the target frame and other frames in the network are also given for each frame. In the case of the \texttt{Experience\_bodily\_harm} frame, it inherits the \texttt{Undergoing} frame, meaning it is a more specific type of event in which an \textsc{Entity} is affected by some \textsc{Event}. It is also presupposed by the \texttt{Cause\_harm} frame, meaning that, for someone to cause a harm on someone else, this last person has to experience a bodily harm. This is represented by the using relation indicated in \ref{ex:frame}. Finally, there is an alternation to our exemplar frame where a stative framing is adopted: the \texttt{Being\_hurt} frame, which is in a inchoative\_of relation with \texttt{Experience\_bodily\_harm}.  

FrameNet domain modeling aims to build a structured cognitive representation of the key concepts within a topic. Each domain comprises frames representing events, participants, and entities connected through frame-to-frame relations. Following \citet{costa2020traducao}, we adapted and expanded this methodology for our study. The next subsections describe the modeling process.

\paragraph{Anonymization and Corpora Compilation}: The first step in modeling the domains was to compile and anonymize a representative corpus for each domain. The open text fields of both e-medical records from e-SUS APS and violence notifications from SINAN were made available by the city of Recife in Brazil.\footnote{e-SUS APS is the system used by the Brazilian Universal Public Healthcare System in primary care units. For the e-medical records, the SOAP notes (Subjective, Objective, Assessment and Plan) were used. SINAN is the National System for Violence Notification.} All open text fields from 3,8 million records were split into 21 million sentences using Trankit, a transformer-based NLP toolkit,  \citep{nguyen-etal-2021-trankit}.

All sentences used in the research underwent a three-step anonymization process: automatic, semiautomatic, and manual validation. The automatic process used three Named Entity Recognition models \citep{souza2019portuguese,pierreguillou2021nerbert,make4010003}, regular expressions, and a fuzzy search based on known venues to detect potential PII. A text span was anonymized only if at least two techniques agreed that it contained sensitive data. The semi-automatic step used frequency lists of Brazilian Portuguese words and common names to identify potential personal names to accelerate manual validation. Lastly, manual validation was done only over specific samples of the corpus. The goals was to evaluate the process and to guarantee full anonymization of sentences that would be made available for the researchers doing frame-based annotation or evaluating semantic patterns.

The corpora compiled for domain modeling comprised 206,000 words distributed into 43,000 sentences for the e-medical records, and 172,000 words distributed into 16,000 sentences for the violence notifications.

\paragraph{Domain analysis and Scenario Frame Creation:} From a sample of the corpora, the study of the domains was conducted to identify the possible participants, actions, places, and objects that could and/or should be present in each of the domains. Based on this study, the first frame for each of the domains could be modeled: the scenario frame. This is a generic frame whose role is to organize the other frames in the domain. In the case of the Health and Violence domains, existing frames were adjusted to occupy this position. The previous frame \texttt{Medical\_interaction\_scenario} was restructured to be broader and became the \texttt{Healthcare\_scenario} frame. Meanwhile, the \texttt{Violence} frame was adjusted to become the \texttt{Violence\_scenario} frame. 

\paragraph{Candidate Term Extraction:} In the next step, the \textit{Keywords –– Terminology extraction} feature of Sketch Engine \citep{kilgarriff2014sketch} was used to extract the terms––single or multiword expressions––most relevant to the domains. This feature calculates a score that highlights the terms that are especially prominent in a focus corpus; thus, the higher the score, the more characteristic or representative the term is of the specific domain. For example, terms such as \textit{abuso.n 'abuse.n'} or \textit{soco.n 'punch.n'} stood out as more characteristic of the Violence domain, compared to terms such as \textit{árvore.n 'tree.n'} or \textit{casa.n 'house.n'}. To accomplish that, the anonymized corpora were compared against a generic corpus, the Portuguese Web 2020. The extracted list of terms consisted of the 2,000 candidates most likely to belong to the domain.

\paragraph{Lexicographic Behavior Analysis:} The lexicographic behavior of the candidate terms in the list was analyzed and potential lexical units (LUs) were grouped together with the aim of identifying those belonging to existing frames or suggesting new frames that needed to be modeled to better structure the domain. For example, a large number of candidate LUs denoted physical assaults such as \textit{machucar.v 'hurt.v'} that could be directly linked to the \texttt{Cause\_harm} frame in the violence domain. However, this frame covers the perspective of the person causing the harm, not that of the victim, which led to the modeling of a new \texttt{Being\_hurt} frame. A considerable number of candidate LUs could also be associated with frames that did not belong to the domains, but that could be of interest to the research objective, such as those that described relations between family members or acquaintances and were linked to the \texttt{Kinship} and \texttt{Personal\_relationship} frames.

\begin{figure*}[!ht]
    \centering
    \includegraphics[width=1\linewidth]{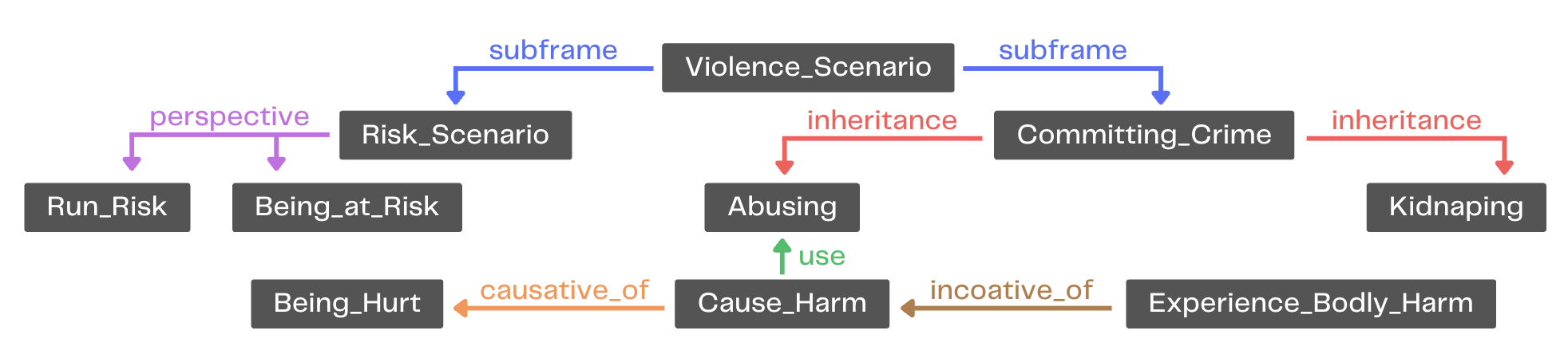}
    \caption{Part of the frames in the Violence domain}
    \label{fig:Violence domain}
\end{figure*}

\paragraph{Frame Creation:} The next step was to model the newly suggested frames. In this process, two approaches were followed: the bottom-up approach, using the study of the lexical items as its staring point, and the top-down approach, where the researcher's intuition is the starting point, as described in \citet{ruppenhofer2016framenet}. For the Healthcare domain, 17 new frames were modeled, six frames were adjusted, and 12 existing frames were linked to the domain, resulting in 35 frames. For the violence domain, five new frames were modeled, one was adjusted, and 42 existing frames were linked, totaling 48 frames.

\paragraph{Frame Relation Definition:} Once the frames for each domain were defined, they were connected through frame-to-frame relations. The relations were defined between frames both in and outside the domains when needed. Following \citeauthor{ruppenhofer2016framenet}, there are eight types of relation between frames: inheritance, subframe, use, causative of, inchoative of, precedence, perspective of and see also. Figure \ref{fig:Violence domain} illustrates the process by showing the relations between some of the frames in the Violence domain.

\paragraph{Lexical Unit Creation:} Once the network of frames is in place, the lemmas grouped as possible LUs were associated with the frames in the database. This step consists of first adding the lemmas and word forms––including not only inflectional, but also spelling variations, abbreviations and acronyms typically found in e-medical records––to the database management and annotation tool \citep{torrent2024flexible}, so that their instantiations in the corpora could be identified in the annotation phase. In total, the Violence domain comprises 1,764 LUs, while the Healthcare domain counts 2,776 LUs. 

\begin{figure}
    \centering
    \includegraphics[width=1\linewidth]{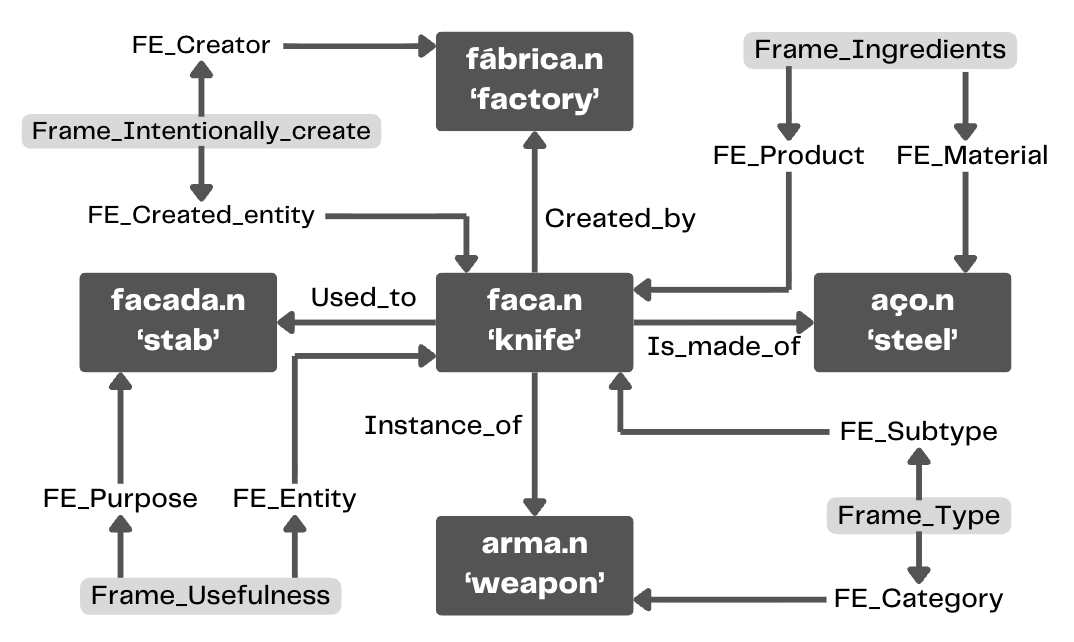}
    \caption{Example of Ternary Qualia Relations}
    \label{fig:TQR}
\end{figure}

\paragraph{Ternary Qualia Relation Definition:} Although frame-to-frame relations play an important role in the model, they are not able to specify which LUs in each frame are to be related to which ones in another frame and also what is the nature of such a relation. To address this issue, Ternary Qualia Relations were defined. These relations are based on the Generative Lexicon Theory \cite{Pustejovsky1995-PUSTGL}, which argues that the meaning of a word can be perspectivized from four types of qualia: agentive, formal, constitutive, and telic. \citet{torrent2024flexible} expanded the theory to have such relations mediated by frames, as illustrated in Figure \ref{fig:TQR} for the LU \textit{faca.n 'knife.n'}. Over 4,500 relations have been created between the LUs associated with the frames in the Healthcare and Violence domains. 

\section{Data Annotation} 
\label{sec:anno}

Annotation was divided into two phases: human and automatic. Given the nature of the data, as an additional layer of protection for the privacy of patients, a version of the annotation platform with restricted access to the team members was created. Anonymized sentences of e-medical records were then uploaded to this instance of the platform to be annotated. The first goal of the manual annotation was to validate the modeling of the domains.  The annotation allows for a general analysis that determines the adequacy of the model, especially with regard to the sufficiency and adequacy of frames in terms of coverage and structure. The second goal of this process was to create a domain-specific training dataset for an automatic semantic role labeler. Both processes are described in detail next.

\subsection{Human Annotation}
\label{sec:human-anno}

The human annotation process was conducted by seven annotators trained in FrameNet methodology \citep{ruppenhofer2016framenet}. The annotation process focused on the semantic content––instead of abiding by the traditional three-layer semantic and syntactic annotation usually employed by FrameNet––which means that the annotation consisted of tagging the frame evoked by the target LU and its frame elements. Also, the expansion of the syntactic locality of the annotation was used in some cases, so as to keep track of core frame elements expressed at the beginning of very long sentences or fragments where automatic sentence segmentation fails, typical of e-medical records, where healthcare professionals may not follow punctuation conventions. The annotation was performed in 2,352 sentences, which yielded 6,313 annotation sets––each of which has one target LU––for the Healthcare domain and 8,309 for the Violence domain. 

\subsection{Automatic Annotation}
\label{automatic-anno}

Automatic annotation used an instance of LOME \citep{xia2021lome}. LOME is a system developed for multilingual information extraction that includes FrameNet semantic parser based on XLM-RoBERTa \citep{conneau-etal-2020-unsupervised}. This parser was re-trained from scratch, having the human annotation described in Section \ref{sec:human-anno} as part of its training dataset in addition to more than 130,000 other annotation sets from previous domain-general annotations available in the FN-Br database. The newly trained parser\footnote{\url{https://github.com/frameNetBrasil/span-finder}} has in its tag set over 200 new frames that expand on FrameNet 1.7 –– the database version used by \citet{xia2021lome}. 


The performance of the LOME instance trained for the project was evaluated and achieved a micro-F1 of 50.68 on the exact matching metric , which is slightly lower than the 56.34 of the original implementation. This reduction in micro-F1 for this metric is considered small given the more challenging nature of the training set:  it contains domain specific, noisy clinical texts in a less resourced language––Brazilian Portuguese––, and a larger set of labels of frames and frame elements.

Before feeding the whole corpus of 21 million sentences to LOME, Trankit was used for tokenization, lemmatization and part-of-speech tagging. The tokenized sentences were fed to LOME and the lemmas, in combination with part-of-speech and the predicted frames were used to infer lexical units, as the FrameNet parser does not predict those out of the box. The whole process resulted in 59 million automatic annotation sets, an average of 2,81 sets per sentence. The computational cost of the full annotation pipeline -- parser retraining (7h51m) and inference over 21 million sentences (62h3m) -- on a single NVIDIA A30 GPU (165W TDP) amounted to approximately 23.06 kWh of energy and an estimated 1.98 kgCO\textsubscript{2}eq, assuming a PUE of 2.0 and a regional grid carbon intensity of 86 gCO\textsubscript{2}eq/kWh for the Centro-Sul region of Brazil \citep{electricitymaps2025}, following the estimation methodology of \citet{lacoste2019quantifying}.

As an elicited example of the resulting automated semantic annotation process, consider the sentence in \ref{ex:lome}, which was annotated by the LOME instance just described for the \texttt{People\_by\_health\_condition}, \texttt{Statement}, \texttt{Symptoms}, \texttt{Body\_parts}, \texttt{Cognitive\_connection}, and \texttt{Emotion\_directed} frames, as shown in (2a-f), respectively. Lexical units evoking each frame are shown in boldface, frame element names are in superscript.

\ex.\label{ex:lome} Patient reports pain in the chest associated with household stress.
\a. [\textbf{Patient}]\textsuperscript{Patient} reports pain in the chest associated with household stress
\b. [Patient]\textsuperscript{Speaker} \textbf{reports} [pain in the chest associated with household stress]\textsuperscript{Message}
\c. [Patient]\textsuperscript{Patient} reports \textbf{pain} [in the chest]\textsuperscript{Body\_part} [associated with household stress]\textsuperscript{Descriptor}
\d. Patient reports pain in the [\textbf{chest}]\textsuperscript{Body\_part} associated with household stress
\e. Patient reports [pain in the chest]\textsuperscript{Concept\_1} \textbf{associated} [with household stress]\textsuperscript{Concept\_2}
\f. Patient reports pain in the chest associated with [household]\textsuperscript{Reason} \textbf{stress}

\section{Construction of Semantic Patterns of GBV}
\label{sec:pattern}


Referring back to the definition of frames, they describe structures that represent contextual meaning and enable the systematic identification of linguistic and semantic patterns that may reveal experiences of violence. Therefore, in order to better understand the scenarios reported in medical records, as well as to look for patterns that could lead to possible unreported and overlooked victims of GBV, an analytical methodology was devised into two phases: \textbf{pattern proposition} and \textbf{validation through corpus querying}.

\subsection{Pattern proposition}
\label{pattern-manual}

The first phase consisted of manually analyzing texts to propose frame-based patterns. It was carried out through the study of the same anonymized sample used in the human annotation that resulted in the dataset used for training LOME. This process was done using two methods.

First, a group of eight possible scenarios covering types of violence was selected to be investigated. They were general violence, psychological violence, sexual abuse, negligence, infant labor, violence against LGBTQIA+ groups, violence against indigenous people and violence against people with disabilities. Subsequently, the LUs that could describe aspects in such cases of violence were selected and grouped to be used in a search for occurrences in the data. Once an LU was identified as present in the record, the report was individually studied to consider: (1) the frame evoked; (2) the frame elements present; (3) the lemmas marked as the frame elements; and (4) the frames those lemmas could be evoking. Based on that analysis, a possible pattern would then be defined.

To illustrate this process, consider the theme of general violence. The lemma \textit{bater.v ‘hit.v’} was chosen as a potential LU capable of indicating instances of such cases. Then this lemma was searched for in the corpus sample, which revealed examples similar to \ref{ex:pattern}. 

\ex.\label{ex:pattern}Paciente relata que o marido bate nela quando fica nervoso.\\ 
\textit{Patient reports that husband hits her when he gets angry.} 

In this case, the LU evokes the \texttt{Cause\_harm} frame, whose core frame elements are \textsc{Agent}––the person causing the Victim's injury––and \textsc{Victim}––the being or entity that is injured. Accordingly, the FE \textsc{Agent} corresponds to \textit{marido.n 'husband'}, while the FE \textsc{Victim} corresponds to \textit{paciente.n 'patient}. The LU \textit{marido.n 'husband'}, in turn, evokes the \texttt{Personal\_relationship} frame. Therefore, we can link the aggressor as someone related to the victim, and a conceptual pattern can be identified: \textit{Physical violence by a family member or person related to the victim}, where the FE \textsc{Agent} in the \texttt{Cause\_harm} frame is filled by an LU that evokes the  \texttt{Personal\_relationship} frame.

A second method to propose patterns consisted of the study of frames in the violence domain. The aim of this method was to further explore the non-core frame elements and how they could also carry relevant information based on the frames the lexical material they labeled were evoking. Each violence-domain frame was analyzed to determine which core and non-core frame elements were relevant when filled by specific lexical units. This allowed for the exclusion of some of the frames given its more generic character. Once the analysis was performed, the sample was again checked.

At the end of both processes, 19 patterns were defined within the eight possible scenarios of violence initially envisioned. This indicates that for some types of violence, more than one pattern was designed. For example, the pattern of \textit{Physical violence by a family member or person related to the victim} was defined under the 'general violence' topic, which is based on the perspective of the agent who causes harm to a victim. However, a second pattern was defined under the same topic, focusing on the victim’s perspective––that of experiencing an injury––activated by the \texttt{Experiencing\_bodily\_harm} frame. The decision to keep a pattern for the next phase was based on the number of occurrences in the sample and/or the researchers’ intuition. 

\subsection{Validation through corpus querying}

The second phase of pattern construction consisted of querying the 19 patterns against the full corpus of 21 million sentences with automatic annotations. Two criteria were established to select useful patterns. For the first, any pattern with less than 30 matches in the corpus had to be discarded. This was the case for six patterns. The second criterion was based on manual inspection of the results (or a sample when they had more than a hundred matches): when patterns had too many false positives, they had to be discarded as they require further refinement and could not be useful as is. This was the case for five additional patterns. 

The remaining eight patterns were kept for analysis and are listed in Section \ref{sec:results}. They cover general violence, sexual violence and negligence. All patterns designed to identify cases of violence against LGBTQIA+, indigenous, and people with disabilities groups had less than ten matches, while queries related to infant labor and psychological violence had too many false positives. In the case of psychological violence, patterns related to fear covered fear of many things that are not necessarily related to violence, while threats were not common. Psychological violence is discussed further in Section \ref{error-analysis}.


\section{Evaluation Design}
\label{sec:experiment}

To evaluate the precision of the proposed methodology in identifying reports of GBV in the e-medical records, the eight specific fine-grained patterns of GBV defined in Section \ref{sec:pattern} were queried against the automatically annotated corpus of sentences and the 5,000 matching sentences extracted. This sample was manually validated one more time to guarantee its full anonymization. Duplicated sentences were removed from the sample, resulting in a total of 4,186 texts.

Human evaluation of the the sample was conducted by six trained annotators, which were familiar with the phenomenon of GBV and with how patterns were defined. Evaluators had to indicate, in the annotation, whether the fragment was an exact instance of the pattern under consideration or not. An exact matching occurs when all parts of the pattern are present in the sentence and the patient is the victim. Partial correspondences with the pattern were judged as a non-matching instances. For example, for the pattern that specifies \textit{injury by family member or person related to the victim}, cases where it was not possible to ensure that an aggressor was identified in the record or that they were not related to or acquainted with the victim were judged as non-matching. Similarly, if the pattern was referencing a person who was not the patient or if it was part of larger pattern such as 'attempt of', it was treated as non-matching. The goal of this round of evaluation is to capture the precision of the methodology in the fine-grained identification of cases of GBV.

Following this first round of evaluation, the same team of annotators performed an error analysis. For this second round, annotators had to indicate whether the mismatches identified in the first round were instances of GBV reports, nevertheless belonging to a different pattern, having partial match or describing, for example, an speculation of violence. In other words, they had to evaluate whether the system correctly identified a case of GBV that would be relevant for a health surveillance team, even if it incorrectly classified its fine-grained type. For this the second round, text fragment batches were shuffled among annotators so that one the same annotator would never judge a fragment they had already analyzed in the first round.

\section{Results}
\label{sec:results}

\subsection{Precision of the Pattern Identification System}

Table \ref{tab:precision} shows the results obtained for the first round of evaluation for the eight patterns of GBV described in Section \ref{sec:pattern}.

\begin{table*}[h]
    \begin{center}
    \begin{tabularx}{\textwidth}{l|r|r|c}
    \hline
    \textbf{Pattern} & \textbf{Correct} & \textbf{Error} & \textbf{Precision} \\
    \hline
    Abandonment of child/elderly & 92 & 27 & 0.773 \\
    Abuse by family member or person related to the victim & 483 & 140 & 0.775 \\
    Abuse of child/elderly & 92 & 17 & 0.844 \\
    Injury by family member or person related to the victim & 493 & 386 & 0.560 \\
    Physical violence by family member or person related to the victim & 574 & 1013 & 0.362 \\
    Sexual violence by family member or person related to the victim & 57 & 43 & 0.570 \\
    Sexual abuse & 362 & 90 & 0.800 \\
    Violence by family member or person related to the victim & 525 & 66 & 0.888 \\
    \hline
    \textbf{Overall} & 2678 & 1782 & 0.600 \\
    \hline
    \end{tabularx}
    \caption{Precision in the fine-grained identification of GBV patterns.}
    \label{tab:precision}
    \end{center}
\end{table*}

In general, patterns that specify both the violence subtype and the aggressor were those with lower precision. There are two relevant explanations for this fact. The first is that violence subtypes are usually evoked by a wide range of verbs, increasing the chances of the semantic parser failing to fully learn the semantic properties of these frames. The second is that complex patterns such as relation to aggressor have more points where it can make a mistake, which reduces its matching chance.



\subsection{Error Analysis}
\label{error-analysis}

The error analysis demonstrated that a considerable part of the non-matching cases, were, in fact, cases of GBV reporting. In most cases, the errors in the first evaluation round were related to a partial correspondence between the pattern and the text fragment being evaluated. Consider, for example, the \textit{physical violence by family member or person related to the victim} pattern, which presented the highest increase in precision post-error analysis, as shown in Table \ref{tab:error_results}. 

\begin{table*}[h]
    \begin{center}
    \begin{tabularx}{\textwidth}{l|r|r|c}
    \hline
    \textbf{Pattern} & \textbf{Correct} & \textbf{Error} & \textbf{Precision} \\ 
    \hline
    Abandonment of child/elderly & 95 & 24 & 0.798 \\
    Abuse by family member or person related to the victim & 511 & 112 & 0.820 \\
    Abuse of child/elderly & 94 & 15 & 0.862 \\
    Injury by family member or person related to the victim & 593 & 286 & 0.675 \\
    Physical violence by family member or person related to the victim & 919 & 669 & 0.579 \\
    Sexual violence by family member or person related to the victim & 76 & 24 & 0.760 \\
    Sexual abuse  & 400 & 52 & 0.885 \\
    Violence by family member or person related to the victim & 553 & 38 & 0.936 \\
    \hline
    \textbf{Overall} & 3241 & 1220 & 0.726 \\ 
    \hline
    \end{tabularx}
    \caption{Precision in the identification of GBV reports by pattern}
    \label{tab:error_results}
    \end{center}
\end{table*}


Among the 345 cases that were classified as non-matching in the first round but had some report GBV, 18 different types of violence were found. These were manually classified after the second round of evaluation to identify possible patterns of errors. Although distributed in many types, 90\% of errors were classified in one of the six most common forms of violence. Table \ref{tab:error_analysis_physical} shows the distribution of these most frequent types found during error analysis and groups the remaining into the 'Other' category.

\begin{table}[h]
    \begin{center}
    \begin{tabular}{l|r|c}
    \hline
    \textbf{Violence type} & \textbf{Count} & \textbf{Percentage} \\ 
    \hline
    Psychological & 110 & 31.9\% \\
    Witnessed & 75 & 21.7\% \\
    Aggressive patient & 42 & 12.2\% \\
    Verbal & 33 & 9.6\% \\
    Abandonment & 25 & 7.2\% \\
    Self-inflicted & 25 & 7.2\% \\
    Other & 35 & 10.2\% \\
    \hline
    \textbf{TOTAL} & 345 & 100\% \\ 
    \hline
    \end{tabular}
    \caption{Distribution of types of violence found in the error analysis performed for the \textit{physical violence by family member or person related to the victim} patter}
    \label{tab:error_analysis_physical}
    \end{center}
\end{table}

These results may indicate that the base frame of this pattern (\texttt{Cause\_harm}) may be capturing more than just physical violence. In many instances, the semantic parser labeled this frame on verbs such as \textit{curse}, \textit{despise} or verbs with modification \textit{attack (verbally)}. This sort of nuance points towards the need of more specific examples on the parser's training set to restrict \texttt{Cause\_harm} annotation to physical examples.

Also note that psychological violence is the most common type in error analysis. As noted in Section \ref{sec:pattern}, the psychological violence patterns initially created were not found in the original sample. The three patterns established to identify psychological violence were:
\begin{enumerate}
    \item The LU \textit{psicológico.a 'psychological'} evoking the \texttt{Domain} frame, when its \textsc{Predicate} FE was filled by an LU evoking the \texttt{Experience\_bodily\_harm}, \texttt{Cause\_harm}, or \texttt{Violence\_scenario} frames.
    \item The LUs \textit{ameaça.n 'threat'} and \textit{ameaçar.v 'to threat'}, evoking the \texttt{Risk\_situation} frame, where the \textsc{Dangerous\_entity} FE was filled by an LU evoking either the \texttt{Kinship} or \texttt{Personal\_relationship} frames.
    \item General LUs evoking the \texttt{Fear} frame, where the \textsc{Stimulus} FE was filled by an LU evoking either the \texttt{Kinship} or \texttt{Personal\_relationship} frames.
\end{enumerate}

An analysis of the 110 cases of psychological violence identified during the error analysis shows interesting reasons why the patterns were not found when querying the corpus or had too many false positives \ref{sec:pattern}. First, among the 110 cases, only one features the LU \textit{psicológico.a 'psychological'} evoking the \texttt{Domain} frame. Second, although 33 cases refer to threats, only in approximately half the cases there is an LU evoking the \texttt{Risk\_situation} frame. For the other half of the cases, the threat is introduced by LUs such as \textit{dizer.v 'say'}, which evoke the \texttt{Statement} frame. The notion of threat is inferred by the nature of the content of the reported speech, which would add sensible complexity to the pattern. A very similar situation occurs with the cases where the psychological violence is based on fear, which total 22 instances. In approximately one third of the cases, there is no explicit LU evoking the \texttt{Fear} frame. 
   
The same analysis showed, however, that almost half the cases of psychological violence––49 instances––are reports of gaslighting, where the patient is the victim. Those reports are very diverse in terms of frames, and also represent a challenge for the pattern identification methodology.

\section{Conclusion and Future Work}

This study presented an innovative methodology for identifying underreporting of notifiable health events, focusing on Gender-Based Violence (GBV) through frame-based modeling and semantic parsing. The results show that this linguistically grounded framework effectively captures a wide range of GBV reports with good precision (average of 0.726).

While the patterns defined in this paper are specific to GBV, the core methodological contribution, \textit{i.e.}, semantic parsing grounded in FrameNet combined with pattern definition over automatic annotations, is generalizable to other health surveillance tasks. This capability rests on two properties of FrameNet. First, beyond the domain-specific healthcare and violence frames modeled by \citet{dutra-et-al-2023} and \citet{larre-torrent-2024}, FrameNet's broad network of general-domain and abstract frames already provides conceptual coverage that other tasks can make use of. Second, the methodology for creating new frames, lexical units, and frame-to-frame relations is well-established and has been applied with methodological rigor across decades of FrameNet development. Extending the system to tasks such as suicide prevention or substance abuse requires defining new patterns — and, only where necessary, new domain-specific frames — within this existing and principled infrastructure.

The same explicit, human-readable structure that makes FrameNet extensible also makes it interpretable. This quality makes the approach particularly suitable for healthcare applications: outputs can be easily understood and validated by professionals, even when the model errs. Such transparency strengthens the system’s reliability and reinforces its practical value in real-world settings.

The methodology also aligns with the realities of Brazil’s public healthcare infrastructure, where limited computational resources often constrain large-scale AI systems. The models employed have a lower carbon footprint and can be deployed on modest infrastructure while maintaining data governance and an ethical implementation. As reported in Section~\ref{automatic-anno}, the entire pipeline -- from parser retraining to inference over 21 million sentences -- consumed approximately 22.4 kWh and emitted an estimated 1.93 kgCO\textsubscript{2}eq, confirming the low environmental cost of deploying task-specific small language models in resource-constrained settings.

Future work will focus on developing frame-based risk models and automating the discovery of semantic patterns to reduce manual effort and enhance scalability.

\section{Ethics and limitations}
\label{sec:ethics}

The research presented in this paper was approved by the Research Ethics Committee of the Federal University of Goiás (CAAE: 64733922.3.0000.5083; Approval number: 6.126.995). 

\paragraph{PII Removal:} The handling of the data used in the research reported in this paper required rigorous attention since, given the nature of violence notifications and e-medical records, the information logged in them is extremely sensitive and has the potential to lead to the identification of the patients. Moreover, some of the patients whose records were used are potential victims of violence and their identification could, apart from violating data privacy regulations, jeopardize their safety. Therefore, to ensure the protection of this information, all members involved in the research signed confidentiality agreements. Also, as described in Section \ref{sec:modeling}, all Personally Identifiable Information (PII) for every patient was removed before the team involved both in data annotation for frames and in evaluation of the performance of the pattern identification system had access to any piece of text obtained from violence notification and/or e-medical records. Moreover, access to the data was restricted even among team members, meaning that only anonymized samples were passed on to researchers.

\paragraph{Annotator's compensation:} All annotators are members of the research team. For the annotation described in Section \ref{sec:human-anno}, annotation tasks were part of the work/research plan of team members, which was compensated through monthly stipends that were at least equivalent to the minimum wage defined by local regulations. For the human evaluation of the pattern recognition task, every annotator is a member of the research team and co-author of the paper. With the exception of one volunteer annotator, all annotators are funded by the project with a monthly stipend at least 38\% higher than the local minimum wage. 

\paragraph{Implications of GBV identification to victims:} Although all data used in this research originate from a historical series ranging from 2016 to 2022, we recognize the potential of this methodology to be implemented as an alert system for healthcare professionals in real time. Such a system may expose victims to additional risks and serious analysis should be made in regards to which ranks of healthcare surveillance professionals should have clearance to access data originated from such an alert system. The research team has conducted extensive research on this issue and has been consulting with specialists in the fields of personal data regulation before any further step towards the use of the methodology as an alert system can be pursued.

\paragraph{Technical limitations:} This methodology was developed as part of a broader project aimed at improving the use of health data to prevent tragic outcomes in Brazil. One of its key objectives is to better estimate underreporting of health-related events. To this end, frame-based modeling, semantic parser training, and the formulation and evaluation of linguistic patterns were conducted using texts in Brazilian Portuguese. Although the work focused on this language, the methodology itself is largely language-agnostic. Since only multilingual language models were employed, it can be applied to other languages with little or no adaptation. Nevertheless, the lack of thorough testing in languages other than Portuguese and the potential bias of frames and lexical units toward Brazilian Portuguese remain a notable limitation.

\section{Acknowledgments}

This work was supported by the Patrick J. McGovern Foundation’s acceleration program, the José Luiz Setúbal Foundation, and the Instituto Galo da Manhã. We also express our gratitude to our partners from the Recife Municipal Health Department — Luciana Caroline, Marcella Abath, Natalia Barros, and Yana Lopes — for their valuable collaboration and continuous support throughout this project. Laís Berno is funded by the Coordination for the Improvement of Higher Education Personnel (CAPES – grant 88887.964418/2024-00). Franciany Campos and Kenneth Brown are funded by the Research Support Foundation of the State of Minas Gerais (FAPEMIG - grant 5.45/2021, scholarship IDs 15366 and 11667, respectively). Tiago Torrent is a grantee of the Brazilian National Council for Scientific and Technological Development (CNPq – grant 311241/2025-5).

\section{Bibliographical References}\label{sec:reference}

\bibliographystyle{lrec2026-natbib}
\bibliography{lrec2026-example,ACL_Anthology_part_aa,ACL_Anthology_part_ab,ACL_Anthology_part_ac}


\end{document}